\documentclass{article}

\usepackage{arxiv}

\usepackage[utf8]{inputenc} % allow utf-8 input
\usepackage[T1]{fontenc}    % use 8-bit T1 fonts
\usepackage{hyperref}       % hyperlinks
\usepackage{url}            % simple URL typesetting
\usepackage{booktabs}       % professional-quality tables
\usepackage{amsfonts}       % blackboard math symbols
\usepackage{nicefrac}       % compact symbols for 1/2, etc.
\usepackage{microtype}      % microtypography

\usepackage{graphicx}
\usepackage{subfigure}
\usepackage{amssymb}
\usepackage{amsmath}
\usepackage{float}
\usepackage[section]{placeins}
\graphicspath{ {./images/} }

\title{A Clustering Method Based on Information Entropy Payload}

\author{
 Shaodong Deng \\
  Xian Mikesi Intelligent Technology Co., Ltd.\\
  Xian, China \\
  \texttt{630237828@qq.com} \\
   \And
  Long Sheng \\
  Xian Mikesi Intelligent Technology Co., Ltd.\\
  Xian, China \\
  \texttt{18191763008@163.com} \\
   \And
  Jiayi Nie \\
  Department of Computer Science\\
  University of Manchester\\
  M13 9PL, United Kingdom \\
  \texttt{jiayi.nie@student.manchester.ac.uk} \\
   \And
  Fuyi Deng \\
  School of Energy and Power Engineering\\
  University of Shanghai For Science and Technology\\
  Shanghai, China \\
  \texttt{2135052501@st.usst.edu.cn} \\  
}

\begin{document}
\maketitle
\begin{abstract}
Existing clustering algorithms such as K-means often need to preset parameters such as the number of categories K, and such parameters may lead to the failure to output objective and consistent clustering results. This paper introduces a clustering method based on the information theory, by which clusters in the clustering result have maximum average information entropy (called entropy payload in this paper). This method can bring the following benefits: firstly, this method does not need to preset any super parameter such as category number or other similar thresholds, secondly, the clustering results have the maximum information expression efficiency. it can be used in image segmentation, object classification, etc., and could be the basis of unsupervised learning.
\end{abstract}

% keywords can be removed

\section{Introduction}
Clustering is to classify similar things into the same cluster and dissimilar things into different clusters. Similarity can be measured in different ways from different perspectives. Objects can be abstracted as data with one or more dimensions, attributes of object correspond to data dimensions. Thus, clustering of objects can be viewed as clustering of data in data set.\\
There are many ways to measure the similarity between data. It is common to treat data as points in space, and measure the similarity by the distance between points, such as Euclidean distance and Manhattan distance.\\
The similarity between data is relative. If the similarity is measured by distance, a threshold value can be used as the parameter to determine whether the data is similar. If the similarity distance between data is less than the threshold value, they are considered to be similar and can be collected into the same data cluster and belong to the same category. Otherwise, it is considered to be dissimilar and collected into different data clusters and belong to different categories.\\
For a certain data set, there will be multiple clustering results due to different thresholds. In order to get a certain result, most existing clustering algorithms require one or more parameters to be given in advance, which directly or indirectly determine the threshold of similarity. For example, in K-means clustering, the number of categories K is required to be given in advance. Giving parameters subjectively often causes confusion, and different results will be produced with different parameters given by different people.\\
How to objectively select the best clustering result from the different clustering results given by different similarity degrees requires an evaluation standard to the clustering results. The clustering results obtained according to this evaluation standard should be aligned with the intuitive feeling of people, and also have the advantage of practical efficiency.\\
This paper considers this problem from the perspective of information theory, gives the best clustering result according to the information expression efficiency, that is, the entropy payload in this paper, and realizes a clustering method without preset parameters.

\section{Entropy Payload}
\label{sec:headings}
\paragraph{Information Entropy of Data Set}
Data could be interpreted as the value of random variables. The data in the set correspond to the points in the uncertainty space of random variables. The space represented by these points constitutes the uncertainty space of random variables. In the clustering results of a data set, a cluster corresponds to a subspace of the uncertainty space approximately. For the clustering results obtained according to different similarities, there are different approximate subspace segmentation methods.\\
Assuming that the data set is S and the total number of data is N, it is divided into n clusters, each cluster has $k_{i}$ elements, and we assume that each cluster is considered to be interior uniform, and the information entropy of the data set is\cite{InformationTheory}:

\begin{equation}
H=-\sum_{i=1}^{n} p_{i}\log_{a} p_{i}
\end{equation}
in which:
\begin{equation}
p_{i} = \frac{K_{i}}{N}
\end{equation}

Information Entropy of the data set reflects the degree of chaos of data diversion. If data is with high consistency, the information entropy is small, if data is with low consistency, the information entropy is big relatively.

\paragraph{Entropy Payload}
 The number of clusters may vary with the threshold of similarity, more clusters will be generated with higher threshold, less clusters will be generated with lower threshold.\\
 If one code is used to represent one cluster, more clusters mean more codes needed, thus the cost of information expression is higher. The information entropy means the benefit of information expression of clustering result, the number of clusters means the cost, so the efficiency of information expression can be measured by the average information entropy of each cluster, which we call entropy payload(EP for short):
\begin{equation}
\begin{aligned}
EP&=\frac{H}{n}\\
&=\frac{-\sum_{i=1}^{n} p_{i}\log_{a} p_{i}}{n}
\end{aligned}
\end{equation}
in which:
\begin{equation}
p_{i} = \frac{K_{i}}{N}
\end{equation}

Entropy payload is the average information entropy carried by each cluster, if we imagine a cluster as a truck which can  carry information to express, and the information entropy of clustering result as the total workload of information to express, the carrying capability of such a truck could be measured by entropy payload(EP for short).\\

For different clustering results based on different degrees of similarity, we choose the clustering result with the maximum entropy payload, so that its information expression efficiency is the highest.

\paragraph{Characteristics of Entropy Payload}

The information entropy of a data set is related to the actual diversion of data, it has different value when different similarity is chosen, but the information entropy value does not change continuously with the change of similarity threshold in much cases.\\
We know if a data set is divided into n clusters, the information entropy may take maximum value when each cluster has equal size, we analysis the characteristics of entropy payload with this simplified case. When data set is equally divided into n parts, $p_{i} = \frac{1}{n}$, the information entropy is:
\begin{equation}
\begin{aligned}
H_{max}&=- n \frac{1}{n} \log_{a}\frac{1}{n}=\log_{a}n\\
EP&=\frac{H_{max}}{n}\\
&=\frac{\log_{a}{n}}{n}
\end{aligned}
\end{equation}

we assume EP is a function f with n as variable, then the maximum entropy formula will be:
\begin{equation}
f(n)=\frac{\log_{a}{n}}{n}
\end{equation}

In order to obtain the extreme point, we differentiate f(n) with respect to n:
\begin{equation}
\begin{aligned}
f^{\prime}(n)&=\frac{\frac{n}{n\ln a} - log_{a}{n}}{n^{2}}=0
\end{aligned}
\end{equation}
then we get:
\begin{equation}
\frac{1}{\ln a} = log_{a}{n}
\end{equation}

so:
\begin{equation}
n = e
\end{equation}
where e is the natural constant, equals to about 2.718.

It means that the function f(n) has maximum value when n=e, irrelevant with base a, the maximum value is 
\begin{equation}
f(e)=\frac{\log_a{e}}{e}
\end{equation}
when the base a is 2, the maximum value is about 0.5307, and when the base a equals e, the maximum value is $\frac{1}{e}$, about 0.3679.
\\
It can be interpreted that when clustering is taken on a data set, the data set is apt to be divided into e clusters equally, in this way, it has the maximum information expression efficiency. 
\\
\\
A series of data points are listed as the following table(logarithmic base is 2):
\begin{table}[H]
 \caption{entropy payload when divided into n parts equally}
  \centering
  \begin{tabular}{lll}
    \toprule
    \cmidrule(r){1-3}
    n     & Entropy payload   & Notes\\
    \midrule
    1.0000 	& 0.0000 \\
    2.0000 	& 0.5000 \\
    2.7183 	& 0.5307 &  n=e,entropy payload has maximum value\\
    3.0000 	& 0.5283 \\
    4.0000 	& 0.5000 \\
    5.0000 	& 0.4644 \\
    6.0000 	& 0.4308 \\
    7.0000 	& 0.4011 \\
    8.0000 	& 0.3750 \\
    9.0000 	& 0.3522 \\
    10.0000 	& 0.3322 \\
    \bottomrule
  \end{tabular}
  \label{tab:table}
\end{table}

The figure is as the following:
\begin{figure}[H] % picture
    \centering
    \includegraphics{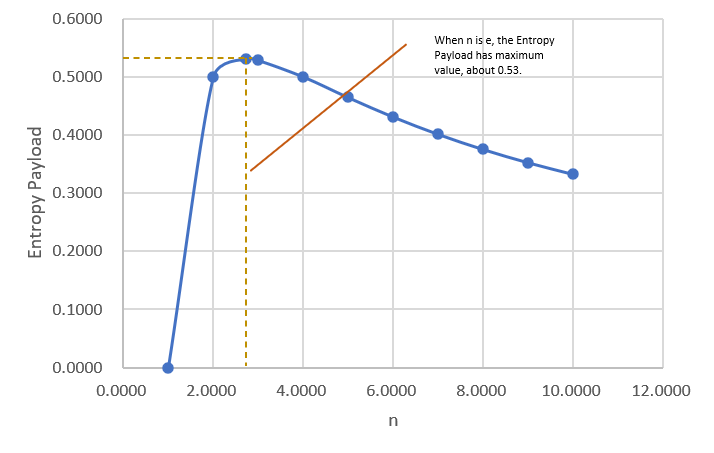}
    \caption{Entropy Payload.}
    \label{fig:fig1}    
\end{figure}

We can know that when n = 1, the data set is regarded as a integral whole, the information entropy is zero, the entropy payload is zero too. \\
When n = e, the entropy payload has the maximum value.
When n increases continuously, the entropy payload will decrease, that means though information entropy increases, but the efficiency of information expression decreases.

\section{Self Entropy Payload}
The entropy payload is the average information entropy payload of cluster, it requires that all data in data set should clustering into clusters according to the same similarity threshold value. \\
If a part of data with an initial similarity is selected as an initial cluster, we can enlarge this data cluster with decreased similarity. We consider the information expression efficiency of this cluster, which can be measured as entropy payload carried by this cluster, we call it self entropy payload(SEP for short), which means the entropy payload carried by the cluster itself, not an average value across multiple clusters.\\
If a cluster with $p_{i}$ as proportion of the whole data set is formed, we define the self entropy payload of the cluster as the entropy payload of clustering result, in which the data set is divided into $\frac{1}{p_{i}}$ clusters equally, so: 

\begin{equation}
\begin{aligned}
EP&=\frac{-\sum_{i=1}^{n} p_{i}\log_{a} p_{i}}{n}
\end{aligned}
\end{equation}
in which n = $\frac{1}{p_{i}}$, and each $p_{i}$ is equal.

then:

\begin{equation}
\begin{aligned}
SEP&=\frac{n p_{i}\log_{a} p_{i}}{n}\\
&=-p_{i}\log_{a} p_{i}
\end{aligned}
\end{equation}

We take SEP as function f with $p_{i}$ as variable:
\begin{equation}
f(p_{i})=-p_{i}\log_{a}{p_{i}}
\end{equation}

In order to obtain the extreme point of $f(p_{i})$, we differentiate f(p$_{i}$) with respect to p$_{i}$:
\begin{equation}
\begin{aligned}
f^{\prime}(p_{i})&=log_{a}{p_{i}}+\frac{1}{\ln a}\\
&=0
\end{aligned}
\end{equation}
then:
\begin{equation}
log_{a}{p_{i}} = - \frac{1}{\ln a}
\end{equation}

so:
\begin{equation}
 p_{i} = \frac{1}{e}
\end{equation}

A series of data points of  $f(p_{i})$ are listed as the following(logarithmic base is 2):
\begin{table}[H]
 \caption{Self entropy payload}
  \centering
  \begin{tabular}{lll}
    \toprule
    \cmidrule(r){1-3}
    Pi     & Self entropy payload   & Notes\\
    \midrule
    0.0001  & 	0.0013 \\
    0.0010  & 	0.0100 \\
    0.0100 	& 0.0664 \\
    0.1000 	& 0.3322 \\
    0.2000 	& 0.4644 \\
    0.3000 	& 0.5211 \\
    0.3679 	& 0.5307    & Pi=1/e,self entropy payload has maximum value\\
    0.4000 	& 0.5288 \\
    0.5000 	& 0.5000 \\
    0.6000 	& 0.4422 \\
    0.7000 	& 0.3602 \\
    0.8000 	& 0.2575 \\
    0.9000 	& 0.1368 \\
    1.0000 	& 0.0000 \\
    \bottomrule
  \end{tabular}
  \label{tab:table}
\end{table}

The figure is as the following:
\begin{figure}[H] % picture
    \centering
    \includegraphics{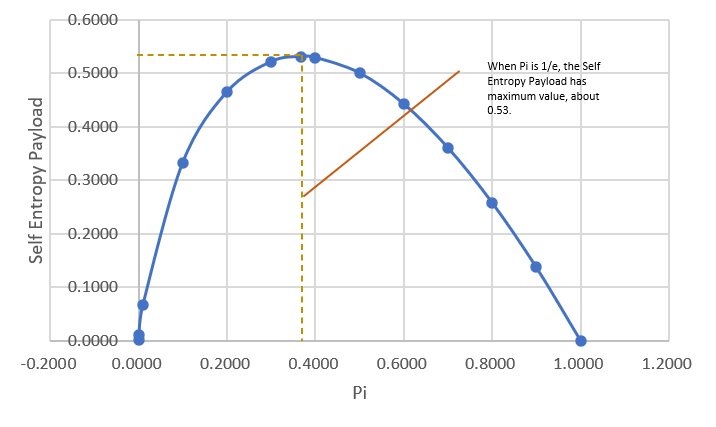}
    \caption{Self Entropy Payload.}
    \label{fig:fig2}    
\end{figure}

We can find that when $p_{i}$ is $\frac{1}{e}$, the SEP has maximum value. The remaining proportion is 1-1/e, it is about 0.632, we know the golden ratio is 0.618, consider the biological brain should have the best efficiency, and this cluster has maximum information expression efficiency, maybe there are some relation between them?

\section{Examples of Clustering Based on Entropy Payload}
We describe the clustering process using a simple example, the input data set includes a series of single-dimension data, as the following, it has 21 numbers:
\begin{equation}
    1,1,1,1,2,2,3,3,3,5,5,5,5,5,5,5,8,8,8,8,8
\end{equation}

We define the difference between number as similarity measurement, then it has the following choices:
\paragraph{a. maximum difference=0}when the maximum difference is zero, the data set will be grouped into the following 5 clusters:
\begin{equation}
    \{1,1,1,1\},\{2,2\},\{3,3,3\},\{5,5,5,5,5,5,5\},\{8,8,8,8,8\}
\end{equation}    
The first cluster has 4 number 1, its probability is 4/21, the second cluster has 2 number 2, its probability is 2/21, and so on.  so the EP is:
\begin{equation}
\begin{aligned}
EP&=-\frac{\sum_{i=1}^{n} p_{i}\log_{2} p_{i}}{n}\\
&=\frac{-\frac{4}{21}\log_{2} \frac{4}{21}-\frac{2}{21}\log_{2} \frac{2}{21}-\frac{3}{21}\log_{2} \frac{3}{21}-\frac{7}{21}\log_{2} \frac{7}{21}-\frac{5}{21}\log_{2} \frac{5}{21}}{5}\\
&=0.44
\end{aligned}
\end{equation}

\paragraph{b. maximum difference=1}when the maximum difference is 1, the data set will be grouped into the following 3 clusters:
\begin{equation}
    \{1,1,1,1,2,2,3,3,3\},\{5,5,5,5,5,5,5\},\{8,8,8,8,8\}
\end{equation}    
The first cluster has 9 numbers, its probability is 9/21, the second cluster has 7 numbers, its probability is 9/21, and so on.  so the EP is:
\begin{equation}
\begin{aligned}
EP&=-\frac{\sum_{i=1}^{n} p_{i}\log_{2} p_{i}}{n}\\
&=\frac{-\frac{9}{21}\log_{2} \frac{9}{21}-\frac{7}{21}\log_{2} \frac{7}{21}-\frac{5}{21}\log_{2} \frac{5}{21}}{5}\\
&=0.52
\end{aligned}
\end{equation}

Similarly, we can obtain the clustering result and calculate the entropy payload with maximum difference=2, and maximum difference=3. If maximum difference=3, the data set is gathered as a cluster, and the entropy payload is zero.
\\
So we can have the following table:
\begin{table}[H]
 \caption{entropy payload with different maximum difference}
  \centering
  \begin{tabular}{lll}
    \toprule
    \cmidrule(r){1-3}
    Maximum Difference     & entropy payload   & Notes\\
    \midrule
    0  & 	0.44 \\
    1  & 	0.52 &  Maximum entropy payload, clustering result\\
    2 	& 0.40 \\
    3 	& 0.0000 \\
    \bottomrule
  \end{tabular}
  \label{tab:table}
\end{table}

From the above table, we can infer that the clustering result with maximum difference with 1 has maximum entropy payload, and the best clustering result is:
\begin{equation}
    \{1,1,1,1,2,2,3,3,3\},\{5,5,5,5,5,5,5\},\{8,8,8,8,8\}
\end{equation}    

\section{Examples of Clustering Based on Self Entropy Payload}
Using the above example, we demonstrate clustering based on self entropy payload.\\
We can select a part of data from data set, now we select the initial part of data, {1,1,1,1}, with the highest similarity, the maximum difference between data in it is 0. 
From the above calculation, we know the SEP
\begin{equation}
\begin{aligned}
SEP&=-\frac{4}{21}\log_{2} \frac{4}{21}\\
&=0.46
\end{aligned}
\end{equation}

Then we lower the similarity by increasing maximum difference between data to 1, a bigger cluster will be formed:  {1,1,1,1,2,2,3,3,3}, the SEP of it is :
\begin{equation}
\begin{aligned}
SEP&=-\frac{9}{21}\log_{2} \frac{9}{21}\\
&=0.52
\end{aligned}
\end{equation}

We continue to lower the similarity by increasing maximum difference between data to 2, a bigger cluster will be formed:  {1,1,1,1,2,2,3,3,3,5,5,5,5,5,5,5}, the SEP of it is :
\begin{equation}
\begin{aligned}
SEP&=-\frac{16}{21}\log_{2} \frac{16}{21}\\
&=0.30
\end{aligned}
\end{equation}

At the last when the maximum difference between data is 3, all data are gathered into one cluster, it's SEP is 0.

The SEPs can be listed in the following table:
\begin{table}[H]
 \caption{self entropy payload}
  \centering
  \begin{tabular}{lll}
    \toprule
    \cmidrule(r){1-3}
    Maximum Difference     & self entropy payload   & Notes\\
    \midrule
    0  & 	0.46 \\
    1  & 	0.52 &  Maximum self entropy payload\\
    2 	& 0.30 \\
    3 	& 0.0000 \\
    \bottomrule
  \end{tabular}
  \label{tab:table}
\end{table}

So we can obtain that from the initial number 1, the cluster with maximum self entropy payload is {1,1,1,1,2,2,3,3,3}.
\\
We can obtain a cluster with maximum self entropy payload from any data. If we begin clustering from number 2 or 3 in the above example, we can get the same cluster as from number 1; if we begin clustering from number 5, we can get the cluster:{5,5,5,5,5,5,5}; if We begin clustering from number 8, we can get the cluster:{8,8,8,8,8}.
\\
When we enlarge the cluster from a initial data, we only concern the data within the current similarity, so the calculation and the clustering can be finished locally, and reduce the calculation compared to clustering based on entropy payload.

\section{Application of Clustering Based on Entropy Payload: Image Segmentation}

As an image, data set includes all pixels, image segmentation can be supported using clustering method. The similarity can be measured by distance between pixels in color space.The following is the clustering result based on maximum entropy payload.The resolution of the original image is 1080*720p, only 1008 sample points are selected evenly to make clustering. \\
The original image is segmented into 5 areas as Figure \ref{fig:fig3}, (a) is the original image. The first area has 572 sample points, which is highlighted as green point in (b), the Second area has 237 sample points as (c), the third area has 111 sample points as (d), and so on:
\begin{figure}[H]
\centering
\subfigure[original image]{
\begin{minipage}[t]{0.4\linewidth}
\centering
\includegraphics[width=1\linewidth]{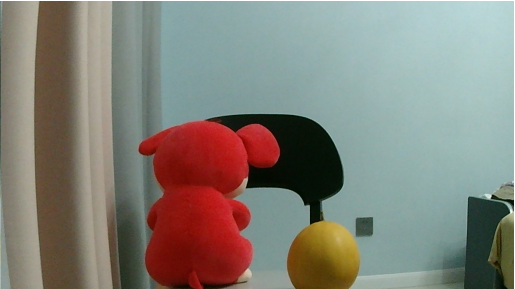}
%\caption{fig1}
\end{minipage}%
}%
\subfigure[area 1:572 sample points]{
\begin{minipage}[t]{0.4\linewidth}
\centering
\includegraphics[width=1\linewidth]{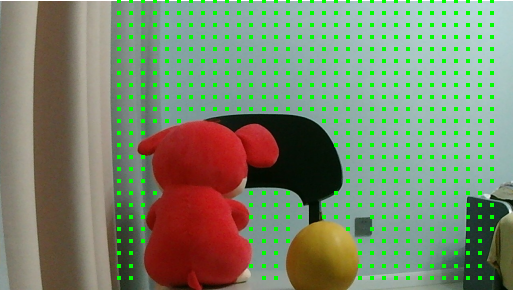}
%\caption{fig2}
\end{minipage}%
}%
\quad
\subfigure[area 2:237 sample points]{
\begin{minipage}[t]{0.4\linewidth}
\centering
\includegraphics[width=1\linewidth]{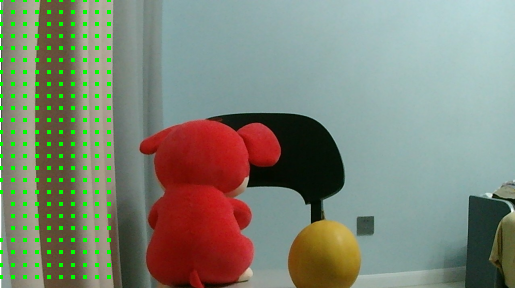}
%\caption{fig2}
\end{minipage}
}%
\subfigure[area 3:111 sample points]{
\begin{minipage}[t]{0.4\linewidth}
\centering
\includegraphics[width=1\linewidth]{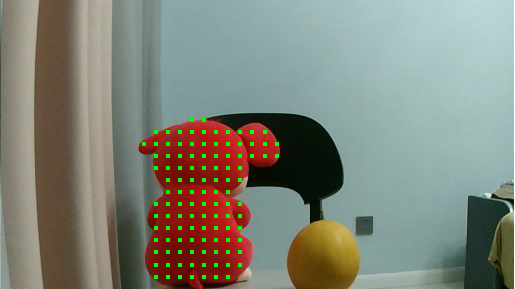}
%\caption{fig2}
\end{minipage}
}%
\quad
\subfigure[area 4:43 sample points]{
\begin{minipage}[t]{0.4\linewidth}
\centering
\includegraphics[width=1\linewidth]{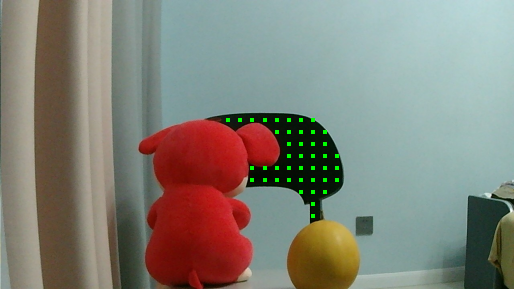}
%\caption{fig2}
\end{minipage}
}%
\subfigure[area 1:25 sample points]{
\begin{minipage}[t]{0.4\linewidth}
\centering
\includegraphics[width=1\linewidth]{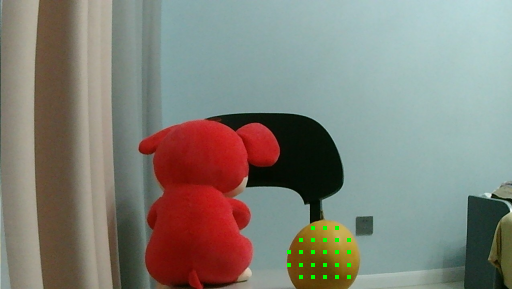}
%\caption{fig2}
\end{minipage}
}%
\centering
\caption{ Image segmentation of original image}
\label{fig:fig3}    
\end{figure}

The area 1 with 572 sample points can further be segmented into 4 sub-areas as Figure  \ref{fig4}:

\begin{figure}[H]
\centering
\subfigure[area 11]{
\begin{minipage}[t]{0.4\linewidth}
\centering
\includegraphics[width=1\linewidth]{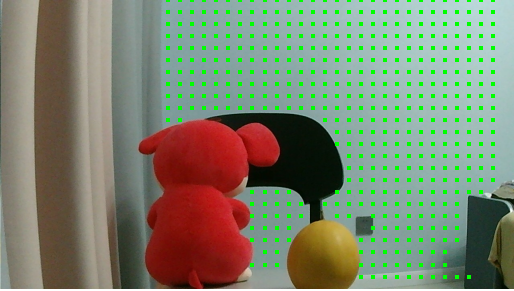}
%\caption{fig1}
\end{minipage}%
}%
\subfigure[area 12]{
\begin{minipage}[t]{0.4\linewidth}
\centering
\includegraphics[width=1\linewidth]{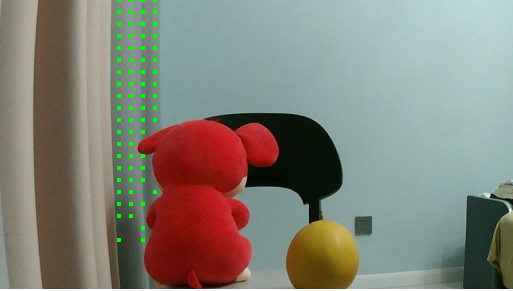}
%\caption{fig2}
\end{minipage}%
}%
\quad
\subfigure[area 13]{
\begin{minipage}[t]{0.4\linewidth}
\centering
\includegraphics[width=1\linewidth]{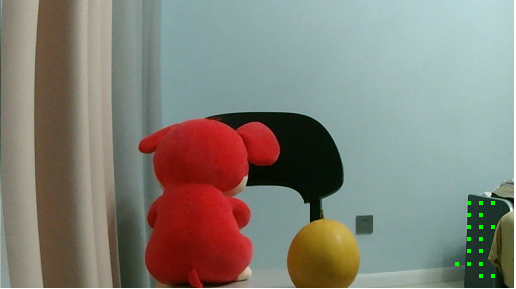}
%\caption{fig2}
\end{minipage}
}%
\subfigure[area 14]{
\begin{minipage}[t]{0.4\linewidth}
\centering
\includegraphics[width=1\linewidth]{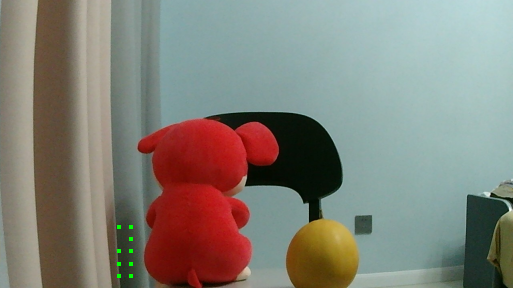}
%\caption{fig2}
\end{minipage}
}%
\centering
\caption{ Image segmentation of area 1 of original image }
\label{fig4}    
\end{figure}

The area 2 with 237 sample points can also further be segmented into 2 sub-areas as Figure  \ref{fig:fig5}:

\begin{figure}[H]
\centering
\subfigure[area 21]{
\begin{minipage}[t]{0.4\linewidth}
\centering
\includegraphics[width=1\linewidth]{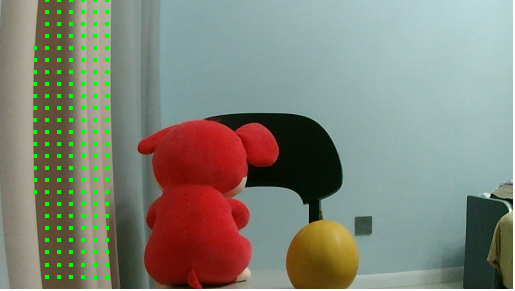}
%\caption{fig1}
\end{minipage}%
}%
\quad
\subfigure[area 22]{
\begin{minipage}[t]{0.4\linewidth}
\centering
\includegraphics[width=1\linewidth]{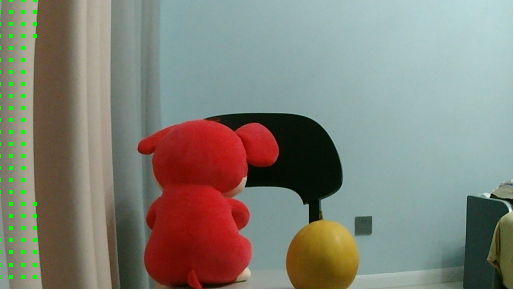}
%\caption{fig2}
\end{minipage}%
}%
\centering
\caption{ Image segmentation of area 2 of original image}
\label{fig:fig5}    
\end{figure}

The hierarchical structure can be shown as Figure  \ref{fig:fig6}, each segmentation is done with the clustering method based on maximum entropy payload on sample points:

\begin{figure}[H] % picture
    \centering
    \includegraphics{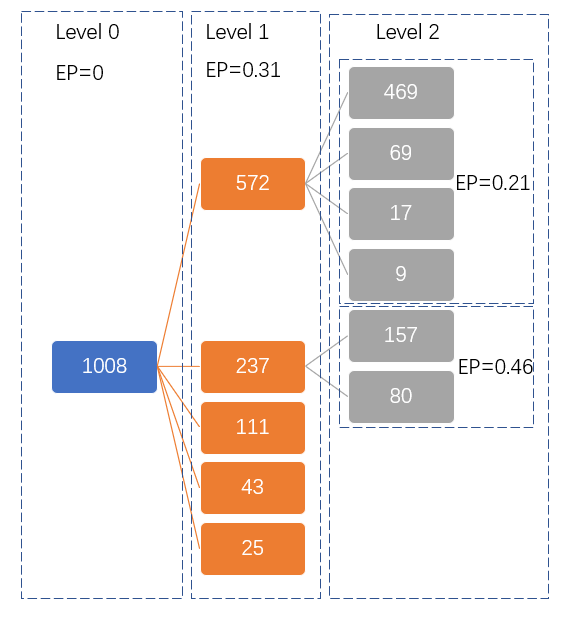}
    \caption{Clustering Hierarchical Structure}
    \label{fig:fig6}    
\end{figure}

The image segmentation can also be supported from a selected sample point at the begin, and enlarge the area including the initial point and calculate the self entropy payload, when the self entropy payload takes the maximum value, the image area can be output as a result area.

\section{Conclusion}
The clustering method based on maximum entropy payload need no preset parameter, so it has the ability to automatic classification, thus provides a way to unsupervised learning. Like the examples of application in image segmentation, visual data of the objective world, which can be extracted from segmented areas, can also further be organized in hierarchical structure in the most efficient way, which can be the basis of object identification in machine vision. \\
The clustering method can not only be used in machine vision, but also can be used in speech recognition. big data analysis, etc. In addition, because it is based on maximum information expression efficiency, it also can be used in encoding of data, such as audio encoding, video encoding, etc., thus maximum efficiency of storage, transfer can be obtained.

\bibliographystyle{unsrt}  
\bibliography{sep}  %%% Remove comment to use the external .bib file (using bibtex).

\end{document}